\documentclass[Journal, 11pt]{IEEEtran}
\ifCLASSINFOpdf
\else
\fi
\usepackage{graphicx}
\usepackage{color}
\usepackage{placeins}
\usepackage{float}
\usepackage{tabularx,colortbl}

\usepackage{amsmath,graphicx}
\usepackage{subfigure}
\usepackage{multirow}
\usepackage{amssymb}
\usepackage{placeins}
\usepackage{color}
\usepackage[ruled]{algorithm2e}

\usepackage{tikz}
\allowdisplaybreaks[1]

\newcommand{\bx}{{\boldsymbol{x}}}
\newcommand{\bu}{{\boldsymbol{u}}}

\newcommand{\bz}{{\boldsymbol{z}}}
\newcommand{\by}{\boldsymbol{y}}

\newcommand{\bhx}{{\boldsymbol{\widehat{x}}}}
\newcommand{\bhz}{{\boldsymbol{\widehat{z}}}}
\newcommand{\bhy}{\boldsymbol{\widehat{y}}}
\newcommand{\bty}{\boldsymbol{\widetilde{y}}}

\makeatletter
\renewcommand{\@algocf@capt@plain}{above}
\makeatother

\DeclareMathOperator*{\argmin}{arg\,min}

\begin{document}
%
\title{Distributed Coding of Multiview Sparse Sources with Joint Recovery}



%
\author{\IEEEauthorblockN{Huynh Van Luong\IEEEauthorrefmark{1},
Nikos Deligiannis\IEEEauthorrefmark{2}, S{\o}ren Forchhammer\IEEEauthorrefmark{3},
and Andr\'{e} Kaup\IEEEauthorrefmark{1}}

\IEEEauthorblockA{\IEEEauthorrefmark{1}Chair of Multimedia Communications and Signal Processing,\\
Friedrich-Alexander-Universit\"{a}t Erlangen-N\"{u}rnberg, 91058 Erlangen, Germany}

\IEEEauthorblockA{\IEEEauthorrefmark{2}Department of Electronics and Informatics, Vrije Universiteit Brussel, 1050 Brussels, Belgium}

\IEEEauthorblockA{\IEEEauthorrefmark{3}DTU Fotonik, Technical University of Denmark, 2800 Lyngby, Denmark}
}


\maketitle

\begin{abstract}
In support of applications involving multiview sources in distributed object recognition using lightweight cameras, we propose a new method for the distributed coding of sparse sources as visual descriptor histograms extracted from multiview images. The problem is challenging due to the computational and energy constraints at each camera as well as the limitations regarding inter-camera communication. Our approach addresses these challenges by exploiting the sparsity of the visual descriptor histograms as well as their intra- and inter-camera correlations. Our method couples distributed source coding of the sparse sources with a new joint recovery algorithm that incorporates multiple side information signals, where prior knowledge (low quality) of all the sparse sources is initially sent to exploit their correlations. Experimental evaluation using the histograms of shift-invariant feature transform (SIFT) descriptors extracted from multiview images shows that our method leads to bit-rate saving of up to 43\% compared to the state-of-the-art distributed compressed sensing method with independent encoding of the sources.

\end{abstract}



%
\IEEEpeerreviewmaketitle

\section{Introduction}
\label{sec:intro}
Recent technological advances in distributed camera networks support emerging application domains such as mobile augmented reality. In this setting, the distributed cameras work collaboratively to achieve a certain computer vision task. In distributed object recognition, for instance, features extracted by the multiview images acquired by the lightweight cameras can be fused to improve the recognition accuracy \cite{YangIEEE10,NaikalFUSION10}. However, distributed smart cameras typically adhere to energy, computational, and bandwidth constraints; furthermore, inter-camera communication should be avoided or kept minimal. Previous works \cite{YangIEEE10,NaikalFUSION10} addressed the aforementioned constraints using schemes based on distributed compressed sensing (DCS) \cite{BaronARXIV09}.


The alternative scheme in \cite{ColucciaEUSIPCO11,ColucciaJCOM14} combined distributed compressive sensing (CS) \cite{BaronARXIV09} with distributed source coding  \cite{dvc:DSlepian,dvc:ADWyner} to reduce the encoding rate and improve the reconstruction of the data. Recently, CS reconstruction with side information (SI) \cite{MotaGLOBALSIP14,MotaARXIV14} was proposed and bounds that predict the number of measurements to reconstruct the data were proposed. In addition, CS was extended to the case where multiple SI signals were used to aid the reconstruction \cite{LuongARXIV16,LuongICIP16}. In a real communication scenario, the schemes in \cite{YangIEEE10,NaikalFUSION10,BaronARXIV09,MotaGLOBALSIP14,MotaARXIV14} do not consider the encoding cost in bit-rate to transmit the measurements: in reality, the measurements at the encoder need to be quantized to a certain bit-depth and encoded efficiently. Meanwhile, the schemes in \cite{ColucciaEUSIPCO11,ColucciaJCOM14} do not deal with the multiple heterogeneous sources with SI. The coding setup \cite{ColucciaEUSIPCO11,ColucciaJCOM14} is based on an asymmetric coding scenario, in which a source and a correlated SI are firstly reduced via the same sensing matrix into their correlated measurements. Thereafter, the source measurement is conditionally decoded given the known SI measurement at the decoder side. In our work, we do not restrict how the coding setup should be exploited and whether sources are conditionally compressed or exploited as SI at the decoder.

We propose an efficient distributed coding of sparse sources (DICOSS) where coarse information as prior information of the sources is initially sent. We consider some reasons why the distributed coding scheme benefits from the prior information: \emph{1)} It can exploit intra-source redundancy given prior information generated from a joint recovering process; \emph{2)} Using all obtained SI signals, it is also able to exploit inter-source correlations among the sources; \emph{3)} It is finally possible to adapt to on-the-fly source changes by deciding the coding set up based on the prior information. These prior information signals are jointly recovered to generate multiple SI signals and then DICOSS employs a distributed source coding for cooperatively decoding. In addition, we propose a joint recovery algorithm incorporating multiple SI signals, which is integrated in DICOSS, to improve the joint recovery.

The rest of this paper is organized as follows. Section \ref{problemBackground} states our problem and reviews the previous works on DCS and distributed source coding. In Sec. \ref{DICOSS}, we present the proposed architecture and we present experimental results on histograms of visual descriptors extracted by multiview images in Sec. \ref{Experiment}. Finally, Section \ref{conclusion} concludes the work.
\section{Problem Statement and Background}
\vspace{-0pt}
\label{problemBackground}
\subsection{Problem Statement}
\label{problem}
\begin{figure*}[!t]
\centering
\setlength{\tabcolsep}{1pt}
\renewcommand{\arraystretch}{0.1}

\subfigure[View 1]{\includegraphics[width=0.17\textwidth]{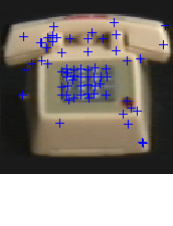}}
\subfigure[$\bx_{1}$]{\includegraphics[width=0.61\textwidth]{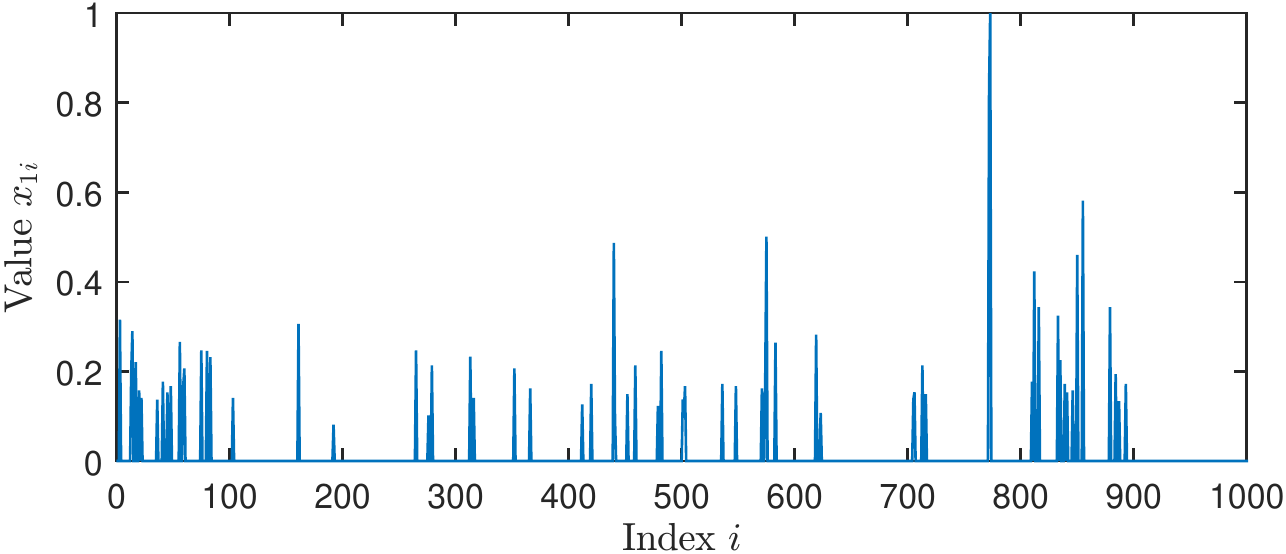}}\\
\subfigure[View 2]{\includegraphics[width=0.17\textwidth]{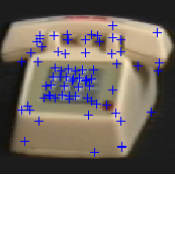}}
\subfigure[$\bx_{2}$]{\includegraphics[width=0.61\textwidth]{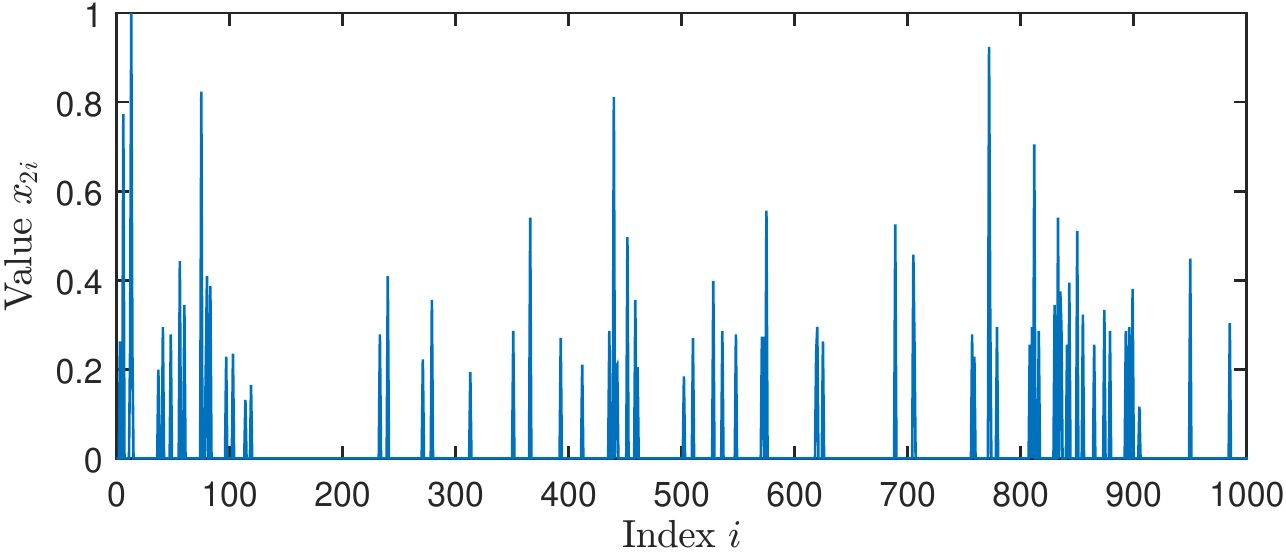}}
\caption{Example of two-view images of Object 60 in COIL-100 \cite{Nene96} with feature points (a) View 1, (b) View 2, and 1000-dimensions (c) $\bx_{1}$, (d) $\bx_{2}$.
}\label{figFeatVec}
\end{figure*}

We consider a problem of how to compress correlated and sparse sources of multiview cameras and transmit them to the decoder for recognizing the object of interest \cite{YangIEEE10,NaikalFUSION10}. Let $\bx_{1}$,...,$\bx_{J}\hspace{-2pt}\in\hspace{-2pt}\mathbb{R}^{n}$ denote $J$ sparse sources, which can represent the corresponding histograms of visual descriptors extracted by $J$ multiview images. Figure \ref{figFeatVec} illustrates two-view images of object 60 in the COIL-100 database \cite{Nene96} with the corresponding SIFT \cite{DLowe99} feature points and correlated histograms $\bx_{1},\bx_{2}$. In the application of distributed object recognition, we may need the high-dimensional histograms under resource-time constraints and the prohibited communication among the lightweight cameras. This arises in a challenge of reducing efficiently the high-dimensional sources before transmitting and then recovering them jointly at the decoder.
\subsection{Background}
\label{relatedWork}
\subsubsection{Distributed Compression for Multiview Sources}
\label{DCforMulti-view}
To solve the distributed compression problem, we consider the basic theory of CS and the joint recovery of DCS \cite{DonohoTIT06,CandesTIT06,BaronARXIV09}. 
CS theory states that a source $\bx\hspace{-2pt}\in\hspace{-2pt}\mathbb{R}^{n}$  can be recovered using the measurement matrix $\mathbf{\Phi}\hspace{-2pt}\in\hspace{-2pt}\mathbb{R}^{m\times n}$ and $m\hspace{-2pt} \ll \hspace{-2pt}n$  linear random measurements $\by\hspace{-2pt}  =\hspace{-2pt}  \mathbf{\Phi}\bx$, where the number of measurements is sufficiently large. Furthermore, DCS \cite{BaronARXIV09} assumes a number of histogram vectors that are each individually sparse and also correlated across the cameras. Each camera independently projects its histogram vector onto an incoherent basis. The decoder can jointly reconstruct each of the signals. DCS utilizes a Joint Sparsity Model (JSM) \cite{BaronARXIV09} to describe both the intra- and inter-camera dependencies. The $J$ sensor signals $\bx_{j}$ can
be written as
\begin{equation}\label{JSM}
    \bx_{j}=\bx_{c}+\bz_{j},
\end{equation}
where the vector $\bx_{c}$ is common to all signals, whereas the vector $\bz_{j}$ is the unique
part of each $\bx_{c}$.
%

Each sparse source $\bx_{j}$ is first reduced by sampling via a linear projection \cite{DonohoTIT06,CandesTIT06}. In particular, we denote a random measurement matrix for $\bx_{j}$ by $\mathbf{\Phi}_{j}\hspace{-2pt}\in \hspace{-2pt}\mathbb{R}^{ m_{j}\times n} (m_{j}\hspace{-2pt}<\hspace{-2pt} n)$, whose elements are sampled from an i.i.d. Gaussian distribution. Thus, on each camera, we get a compressed vector $\by_{j}\hspace{-2pt}=\hspace{-2pt}\mathbf{\Phi}_{j}\bx_{j}$, also called measurement, consisting of $m_{j}$ elements. At the decoder, the ensemble $\bx_{1},\bx_{2},...,\bx_{J}$ can be recovered individually \cite{CandesTIT06} by solving:
\begin{equation}\label{l1-norm}
    \min_{\bx_{j}} \parallel\hspace{-2pt}\bx_{j}\hspace{-2pt}\parallel_{1} \mathrm{subject~to~} \by_{j}=\mathbf{\Phi}_{j}\bx_{j}.
\end{equation}

Furthermore, they are able to be jointly solved in a single linear system by JSM \cite{BaronARXIV09}:
\begin{equation}\label{JSM1}
\left[
      \begin{array}{c}
        \by_{1} \\
        \by_{2} \\
        \vdots \\
        \by_{J} \\
      \end{array}
\right]=
\underbrace{\left[
  \begin{array}{ccccc}
    \mathbf{\Phi}_{1} & \mathbf{\Phi}_{1} & 0 & \ldots & 0\\
    \mathbf{\Phi}_{2} & 0 & \mathbf{\Phi}_{2} & \ldots & 0\\
    \vdots   & \vdots & \vdots & \ddots & \vdots\\
    \mathbf{\Phi}_{J} & 0 & 0 & \ldots &\mathbf{\Phi}_{J}\\
  \end{array}
\right]}_{\mathbf{\Phi}'}
\left[
      \begin{array}{c}
        \bx_{c} \\
        \bz_{1} \\
        \vdots \\
        \bz_{J} \\
      \end{array}
\right].
\end{equation}
When the multiview recovered histograms $\bhx_{j}$ are available, a multiview object recognition \cite{NaikalFUSION10} using a hierarchical vocabulary tree \cite{NisterCVPR06} takes the multiview histograms as the input and outputs a label for the considered object.
\subsubsection{Distributed Source Coding of Sparse Sources}
\label{DSCforSS}
To consider the encoding cost of the measurements $\by_{j}$ \eqref{JSM1} in bits rather than real coefficients, the authors in \cite{ColucciaEUSIPCO11,ColucciaJCOM14} construct a quantized DCS architecture to exploit knowledge of the SI at the decoder. Source coding with SI at the decoder is considered in the Slepian-Wolf (SW) framework \cite{dvc:DSlepian} for lossless distributed coding and Wyner-Ziv \cite{dvc:ADWyner} for lossy distributed coding. The theorems show that two given i.i.d. sources $Y_{1}$ and $Y_{2}$ can be jointly recovered with vanishing error probability when they are encoded separately and decoded jointly with total rate $R_{1}\hspace{-2pt}+\hspace{-2pt}R_{2}\hspace{-2pt}=\hspace{-2pt}H(Y_{1},Y_{2})$ as the joint entropy of $Y_{1}$ and $Y_{2}$. The individual rates of $Y_{1}$ and $Y_{2}$ need to only satisfy $R_{1}\hspace{-2pt}\geq \hspace{-2pt}H(Y_{1}|Y_{2})$ and $R_{2}\hspace{-2pt}\geq \hspace{-2pt} H(Y_{2}|Y_{1})$, where $H(Y_{1}|Y_{2})$ and $H(Y_{2}|Y_{1})$ are conditional entropies.

Let us consider two measurements, $\by_{1}$ and $\by_{2}$, which are sampled from two sparse sources, $\bx_{1}$ and $\bx_{2}$. The coding diagram in \cite{ColucciaEUSIPCO11} imposes that $\bx_{1}$ and $\bx_{2}$ have the same dimensions of the data using the same sensing matrix $\mathbf{\Phi}$. Consequently, after uniformly quantizing, their quantized $\bhy_{1}$ and $\bhy_{2}$ are supposed to be still correlated since $\bx_{1}$ and $\bx_{2}$ are correlated. The system employs the asymmetric setup, where $\bhy_{2}$ is entropy coded and exploited as SI at the decoder and $\bhy_{1}$ is coded by the SW coder. Finally, the reconstruction is performed, in which exploiting SI is also taken into account assuming that the difference of $\by_{1}$ and $\by_{2}$ is modeled as Gaussian additive correlation noise.

\section{DIstributed COding of Sparse Sources (DICOSS) with Joint Recovery}
\label{DICOSS}
We propose a novel method to perform a distributed coding of sparse sources, called DICOSS, by exploiting intra- and inter-source correlation at a central decoder. Our approach is motivated by a distributed object recognition task in which the involved cameras have limitations in terms of computation power and communication bandwidth. Our design is shown to outperform alternatively schemes including the state-of-the-art DCS scheme \cite{YangIEEE10,NaikalFUSION10}. 
Figure 2 presents the proposed DICOSS architecture, which combines a compressed sensing with multiple side information signals \cite{LuongARXIV16,LuongICIP16} with a multiterminal source coding scheme based on asymmetric SW coding scenarios \cite{dvc:DSlepian,dvc:ADWyner}.
\begin{figure*}[!hbt]
  \centering
  \includegraphics[width=0.89\textwidth]{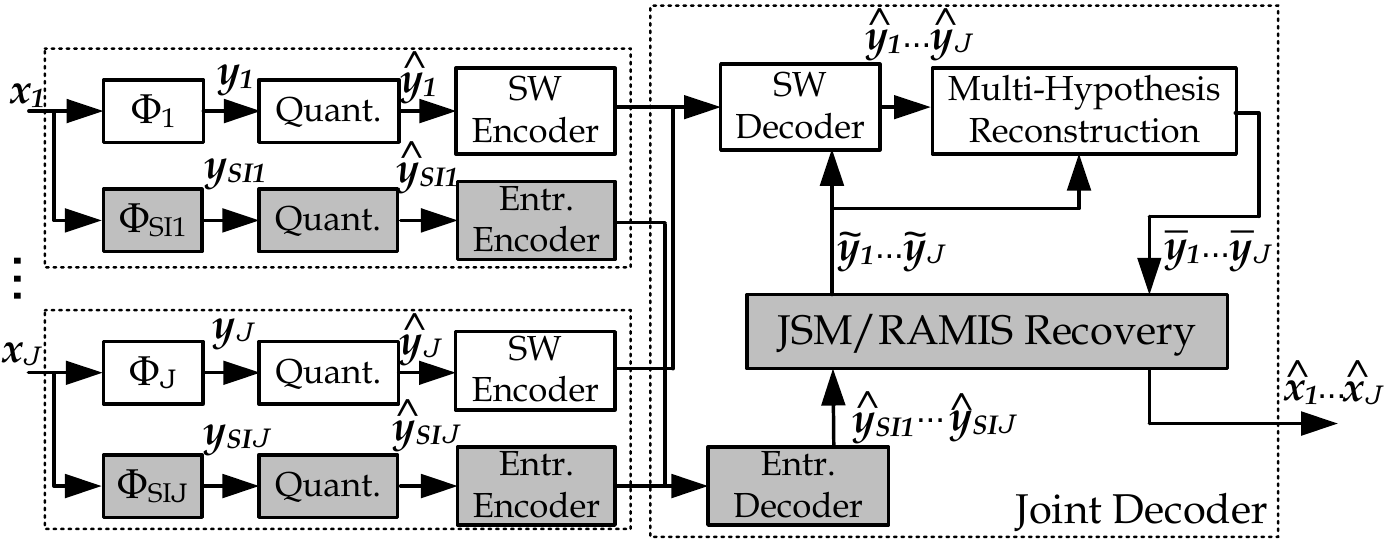}
  \caption{The proposed DICOSS architecture.}
  \label{figDICOSS}
\end{figure*}

At each distributed encoder, we acquire a low and a high resolution measurement vectors of each signal, each denoted as $\by_{j}$ and $\by_{SIj}$, respectively. These measurements are acquired with the corresponding matrices $\mathbf{\Phi}_{j}$ and $\mathbf{\Phi}_{SIj}$. After quantization, the low resolution measurements $\bhy_{SIj}$ are entropy encoded while the high resolution measurements $\bhy_{j}$ are encoded using SW coding (realised using the LDPCA code in \cite{dvc:DVarodayan}). At the decoder, the low resolution measurements are first entropy decoded and jointly used to produce high-quality SI $\bty_{j}$ by a JSM/RAMIS recovery. Thereafter, $\bty_{j}$ are used to decode $\bhy_{j}$ and then the multi-hypothesis reconstruction is applied to dequantize them. Finally, the measurements $\overline{\by}_{j}$ are used in the JSM/RAMIS module to obtain the signals of interest $\bhx_{j}$. We will consider two joint recovery methods as detailed in Sec. \ref{SI_DCS} and the joint SW decoding and multi-hypothesis reconstruction in Sec. \ref{JointDecodeRecconstruct}. It can be noted that all grey blocks in Fig. \ref{figDICOSS} are related to SI.

\subsection{Joint Sparse Signal Recovery}
\label{SI_DCS}
We introduce two joint recovery methods. The first method is to use the joint sparsity recovery model to recover all $\bx_{j}$ by means of the method in DCS \cite{BaronARXIV09}. The second method exploits our proposed RAMSIA algorithm \cite{LuongICIP16} to recover in turn each source $\bx_{j}$ given already reconstructed sources $\bx_{1},...,\bx_{j-1}$.
\subsubsection{Joint Sparsity Model}
\label{JSM_Model}
The purpose of sending coarse information is to jointly generate SI signals at the decoder shown by the highlighted blocks in Fig. \ref{figDICOSS}. Using random projections \cite{CandesTIT06} $\mathbf{\Phi}_{SIj}\hspace{-2pt}\in \hspace{-2pt}\mathbb{R}^{ m_{SIj}\times n} (m_{SIj}\hspace{-2pt}< \hspace{-2pt}n)$, $\by_{SIj}\hspace{-2pt}=\hspace{-2pt}\mathbf{\Phi}_{SIj}\bx_{j}$. After quantization, entropy encoding, and decoding, we obtain $\bhy_{SI1},...,\bhy_{SIJ}$ at the joint decoder. Let $\bx'\hspace{-2pt}=\hspace{-2pt}[\bx_{c}^{T},\bz_{1}^{T},...,\bz_{J}^{T}]^{T}$, $\bhy'_{SI}\hspace{-2pt}=\hspace{-2pt}[\bhy_{SI1}^{T},\bhy_{SI2}^{T},...,\bhy_{SIJ}^{T}]^{T}$, and $\mathbf{\Phi}'_{SI}$ is formulated from $J$ projections $\mathbf{\Phi}_{SIj}$ similar to $\mathbf{\Phi}'$ in \eqref{JSM1}. JSM recovers $\bx'$ based on \eqref{l1-norm},\eqref{JSM1} by solving
\begin{equation}\label{l1-norm-JSM}
    \min_{\bx'} \parallel\bx'\parallel_{1} \mathrm{subject~to~} \bhy'_{SI}=\mathbf{\Phi}'_{SI}\bx'.
\end{equation}
Then we obtain $\bty_{j}\hspace{-2pt}=\hspace{-2pt}\mathbf{\Phi}_{j}\bx_{j}$, where $\bx_{j}$ are derived by \eqref{JSM}. 
Furthermore, after the multi-hypothesis reconstruction, given the reconstructed $\overline{\by}_{j}$, we use JSM/RAMIS (see Fig. \ref{figDICOSS}) to jointly recover $\bhx_{j}\hspace{-2pt}=\hspace{-2pt}\bhx_{c}\hspace{-2pt}+\hspace{-2pt}\bhz_{j}$ by solving
\begin{equation}\label{l1-norm-JSM-reconstruct}
    \min_{\bhx'} \parallel\bhx'\parallel_{1} \mathrm{subject~to~} \by'=\mathbf{\Phi}'\bhx',
\end{equation}
where $\bhx'\hspace{-2pt}=\hspace{-2pt}[\bhx_{c}^{T},\bhz_{1}^{T},...,\bhz_{J}^{T}]^{T}$, $\by'\hspace{-2pt}=\hspace{-2pt}[\overline{\by}_{1}^{T},\overline{\by}_{2}^{T},...,\overline{\by}_{J}^{T}]^{T}$, and $\mathbf{\Phi}'$ is the matrix denoted in \eqref{JSM1}.
\subsubsection{Sparse Signal Reconstruction with Multiple SI Signals}
\label{RAMIS_Model}
An alternative joint recovery method is that we can recover in turn each source $\bx_{j}$ given the reduced $\by_{j}$ and other already reconstructed $\bx_{1},...,\bx_{j-1}$. In order to do this, we propose an Reconstruction Algorithm with Multiple Incremental SI, called RAMIS, by modifying the RAMSIA algorithm \cite{LuongICIP16}, which reconstructs a sparse signal with multiple side information signals. RAMIS is to recover each source $\bx_{j}$ given already reconstructed $\{\bx_{1},...,\bx_{j-1}\}$ as multiple SI signals then $\bx_{j}$ is acquired to increase the previous SI set to $\{\bx_{1},...,\bx_{j}\}$ for the next recovery of $\bx_{j+1}$.

The objective function of RAMIS shall be created based on RAMSIA in \cite{LuongICIP16} as an $n$-$\ell_{1}$ minimization problem of finding a solution to
\begin{equation}\label{l1-general}
    \min_{\bx}\{H(\bx) = f(\bx) + g(\bx)\},
\end{equation}
where $f(\bx)\hspace{-2pt}=\hspace{-2pt}\frac{1}{2}||\mathbf{\Phi}\bx\hspace{-2pt}-\hspace{-2pt}\by||^{2}_{2}$ and $\lambda\hspace{-2pt}>\hspace{-2pt}0$ is a regularization parameter. And the function $g(\bx)$ is defined by
\begin{equation}\label{n-l1g}
g_{j}(\bx_{j}) \hspace{-2pt}= \lambda \hspace{-2pt}\sum\limits_{p=0}^{j-1}\hspace{-2pt}\beta_{p}||\mathbf{W}_{p}(\bx_{j}-\bx_{p})||_{1},
\end{equation}
where $\beta_{p}\hspace{-2pt}>\hspace{-2pt}0$ are weights across SI signals and $\mathbf{W}_{p}$ is a diagonal matrix with weights that correspond to the SI signal $\bx_{p}$, $\mathbf{W}_{p}\hspace{-2pt}=\hspace{-2pt}\mathrm{diag}(w_{p1},...,w_{pn})$, wherein $w_{pi}\hspace{-2pt}>\hspace{-2pt}0$ is the weight in $\mathbf{W}_{p}$ at index $i$ for the given $\bx_{p}$. In particular, for $p\hspace{-2pt}=\hspace{-2pt}0$, $\bx_{p}\hspace{-2pt}=\hspace{-2pt}\mathbf{0}$. We compute weights in two levels, first $w_{pi}$ for intra-SI weights and then inter-SI weights $\beta_{p}$. 
Namely, the objective function of RAMIS by:
\begin{equation}\label{n-l1minimizationGlobal}
    \min_{\bx_{j}}\hspace{-2pt}\Big\{\hspace{-2pt}H(\bx_{j})\hspace{-2pt}=\hspace{-2pt}\frac{1}{2}||\mathbf{\Phi}_{j}\bx_{j}\hspace{-1pt}-\hspace{-1pt}\by_{j}||^{2}_{2} \hspace{-2pt}+\hspace{-2pt} \lambda \hspace{-2pt}\sum\limits_{p=0}^{j-1}\hspace{-2pt}\beta_{p}||\mathbf{W}_{p}(\bx_{j}\hspace{-1pt}-\hspace{-1pt}\bx_{p})||_{1}\hspace{-2pt}\Big\}.
\end{equation}

The proposed RAMIS is described in Algorithm \ref{RAMSIAAlg-IC}. Contrary to RAMSIA \cite{LuongICIP16}, RAMIS here computes in turn $J$ runs of RAMSIA, where it also updates the multiple SI set after each run of recovering $\bx_{j}$. It can be noted that the function $\Gamma_{\hspace{-2pt}\frac{1}{L}g_{j}}(.)$ and \textit{Stopping criteria} in Algorithm \ref{RAMSIAAlg-IC} are defined as in RAMSIA \cite{LuongICIP16}.
\setlength{\textfloatsep}{0pt}
\begin{algorithm}[t!]
\DontPrintSemicolon \SetAlgoLined
\textbf{Input}: $\by_{1},...,\by_{J},\mathbf{\Phi}_{1},...,\mathbf{\Phi}_{J}$;\\
\textbf{Output}: $\bx_{1},\bx_{2},...,\bx_{J}$;\\
   \tcp{\hspace{-2pt}Recovering $\bx_{j}$ given $\bx_{1},...,\bx_{j-1}$.}
 \For{$j=1$ to $J$}{
   \tcp{Initialization.}
    $\mathbf{W}_{0}^{(1)}\hspace{-2pt}=\hspace{-2pt}\mathbf{I}$; $\beta_{0}^{(1)}\hspace{-2pt}=\hspace{-2pt}1$; $\mathbf{W}_{p}^{(1)}\hspace{-2pt}=\hspace{-2pt}\mathbf{0}$; $\beta_{p}^{(1)}\hspace{-2pt}=\hspace{-2pt}0~(1\hspace{-2pt}\leq \hspace{-2pt} p\hspace{-2pt}\leq \hspace{-2pt}j\hspace{-2pt}-\hspace{-2pt}1)$;
   $\bu^{(1)}\hspace{-2pt}=\hspace{-2pt}\bx_{j}^{(0)}\hspace{-2pt}=\hspace{-2pt}\mathbf{0}$; $L\hspace{-2pt}=\hspace{-2pt}L_{\nabla f}$; $\lambda,\epsilon \hspace{-2pt}>\hspace{-2pt}0$; $t_{1}\hspace{-2pt}=\hspace{-2pt}1$; $k\hspace{-2pt}=\hspace{-2pt}0$; \\
  \While{Stopping criterion is false}{
       $k=k+1$; \\
       \tcp{Solving given the weights.}
        $\nabla f(\bu^{(k)})=\mathbf{\Phi}_{j}^{\mathrm{T}}(\mathbf{\Phi}_{j} \bu^{(k)}-\by_{j})$; \\
	   $\bx_{j}^{(k)}\hspace{-2pt}= \hspace{-2pt}\Gamma_{\hspace{-2pt}\frac{1}{L}g_{j}}\hspace{-2pt}\Big(\bu^{(k)}\hspace{-2pt}-\hspace{-2pt}\frac{1}{L}\nabla f(\bu^{(k)})\hspace{-2pt}\Big)$;\\ 
	   \tcp{Computing the updated weights.}
	   $w_{pi}^{(k+1)} = \frac{n}{1+\Big(|x_{ji}^{(k)}-x_{pi}|+\epsilon\Big)\Big(\sum\limits_{l\neq i}(|x_{ji}^{(k)}-x_{pl}|+\epsilon)^{-1}\Big)}$;
\\
   $\vspace{-2pt}\beta_{p}^{(k+1)}\hspace{-2pt} =\hspace{-2pt}$\\ $\vspace{-4pt}\frac{1}{\hspace{-2pt}1\hspace{-2pt}+\hspace{-2pt}\Big(\hspace{-2pt}||\mathbf{W}_{p}^{(k+1)}\hspace{-1pt}(\bx_{j}^{(k)}\hspace{-2pt}-\bx_{p})||_{1}\hspace{-1pt}+\hspace{-1pt}\epsilon\hspace{-2pt}\Big)\hspace{-4pt}\Big(\hspace{-3pt}\sum\limits_{l\neq p}\hspace{-2pt}(||\mathbf{W}_{l}^{(k+1)}\hspace{-1pt}(\bx_{j}^{(k)}\hspace{-2pt}-\bx_{l})||_{1}\hspace{-1pt}+\hspace{-1pt}\epsilon)^{-1}\hspace{-2pt}\Big)\hspace{-2pt}}$;
\\
       \tcp{Updating new values.}
        $t_{k+1}\hspace{-2pt}=\hspace{-2pt}(1\hspace{-2pt}+\hspace{-2pt}\sqrt{1\hspace{-2pt}+\hspace{-2pt}4t_{k}^{2}})/2$;\\
        $\bu^{(k+1)}\hspace{-2pt}=\hspace{-2pt}\bx_{j}^{(k)}\hspace{-2pt}+\hspace{-2pt}\frac{t_{k}\hspace{-2pt}-\hspace{-1pt}1}{t_{k+1}}(\bx_{j}^{(k)}\hspace{-2pt}-\hspace{-2pt}\bx_{j}^{(k-1)})$;\\
  }
  \Return $\bx_{j}^{(k)}$;
  }

\caption{The proposed RAMIS algorithm.}\label{RAMSIAAlg-IC}
\end{algorithm}
\subsection{Joint Decoding and Multi-Hypothesis Reconstruction}
\label{JointDecodeRecconstruct}
The DICOSS architecture has the advantage of yielding multiple SI signals generated by the JSM/RAMIS recovery (Sec. \ref{SI_DCS}). These SI signals are valuable information not only for the SW decoding (Sec. \ref{JointDecode}) but also for the reconstruction process (Sec. \ref{JointReconstruct}).
\subsubsection{Joint Decoding}
\label{JointDecode}
To decode the quantized vectors $\bhy_{j}$, we employ SW coding \cite{dvc:DSlepian}, where the LDPCA code \cite{dvc:DVarodayan} with multiple SI signals \cite{dvc:HVLuongTIP12} is used. The coding efficiency of the LDPCA decoder critically depends on the quality of SI signals and the residual statistics or the noise model between the sources, $\by_{j}$, and the SI signals, $\bty_{j}$. It is worth emphasizing that the strategy of using the Laplacian noise model yields the best results as proved in \cite{MotaARXIV14} rather than using the Gaussian correlation noise model in \cite{ColucciaEUSIPCO11}. Therefore, the residue is here modeled by a Laplacian distribution in this work. Let $y_{j}$, $\widetilde{y}_{j}$ denote corresponding elements of $\by_{j}$, $\bty_{j}$ and the Laplacian distribution is represented by a conditional probability density function of $\by_{j}$ given an element $\widetilde{y}_{j}$ as
\begin{equation}\label{LaplacianPDF}
    f_{\by_{j}|\widetilde{y}_{j}}(y_{j})=(\alpha_{j}/{2})e^{-\alpha_{j}|y_{j}-\widetilde{y}_{j}|},
\end{equation}
where $\alpha_{j}$ is the model parameter related to the variance $\sigma_{j}^{2}$ of the Laplacian distribution by $\sigma_{j}^{2}\hspace{-2pt}=\hspace{-2pt}2/\alpha_{j}^{2}$.

Taking multiple SI signals $\bty_{1},...,\bty_{J}$ into account, we combine the individual distributions into the weighted distribution by:
\begin{equation}\label{weightedLaplacianPDF}
    f_{\by_{j}|\widetilde{y}_{1},...,\widetilde{y}_{J}}(y_{j})=\sum\limits_{j=1}^{J}u_{j}f_{\by_{j}|\widetilde{y}_{j}}(y_{j}),
\end{equation}
where $u_{j}$ denotes a weight on the SI $\bty_{j}$ with  $u_{j}\hspace{-2pt}\geq \hspace{-2pt}0$ and $\sum_{j=1}^{J}u_{j}\hspace{-2pt}=\hspace{-2pt}1$. More specially, strategies of varying the parameters $u_{1},...,u_{J}$ can give different inputs for the LDPCA decoder \cite{dvc:DVarodayan} as well as the adaptive coding setups based on the correlations among sources \cite{Nikos14}. Eventually, the LDPCA decoder \cite{dvc:DVarodayan,dvc:HVLuongTIP12} uses the best soft-input among the multiple-inputs for successfully decoding $\bhy_{1},...,\bhy_{J}$ representing the decoded values within the quantization interval.
\vspace{-0pt}
\subsubsection{Multi-Hypothesis Reconstruction}
\label{JointReconstruct}
The reconstruction is to reconstruct $\by_{1},...,\by_{J}$ and their outputs, $\overline{\by}_{1},...,\overline{\by}_{J}$ to be used to recover $\bhx_{1},...,\bhx_{J}$ as the final results. The $\by_{j}$ are reconstructed based on the multiple SI signals $\bty_{j}$ (Sec. \ref{SI_DCS}), the decoded $\bhy_{j}$, and the noise distributions $\alpha_{j}$ (Sec. \ref{JointDecode}), denoted by $\overline{\by}_{j}$. We can reconstruct elements $y_{j}$ of $\by_{j}$ by applying the reconstruction with multiple SI signals in \cite{dvc:HVLuongTIP14} by:
\begin{equation}\label{weightedReconstruction}
    \overline{y}_{j}=\frac{\sum\limits_{j=1}^{J}u_{j}\int\limits_{L}^{U}y_{j} f_{\by_{j}|\widetilde{y}_{j}}(y_{j})\mathrm{d}y_{j}}{\sum\limits_{j=1}^{J}u_{j}\int\limits_{L}^{U}f_{\by_{j}|\widetilde{y}_{j}}(y_{j})\mathrm{d}y_{j}},
\end{equation}
where $[L,U)$ is the decoded quantization interval of $y_{j}$ in $\bhy_{j}$ and $u_{j}$ is determined from the joint decoding (Sec. \ref{JointDecode}). 
\subsection{Adaptive Rate Allocation}
\label{AdaptiveRate}
An important question is: Is there any rate penalty incurring by DICOSS sending $\by_{SIj}$ plus additional bits for SW coding rather than only sending the original $\by_{j}$? Depending on the intra- and inter-source correlations, there may be a chance that the proposed approach performs worse than sending only $\by_{j}$. In the following, we determine the cases in which either a direct encoding of $\by_{j}$ or the proposed strategy is preferable (we refer to the former and the latter as Intra- and Prior-mode, respectively). 
As a result, we still have a generalized scheme which is transparent to the specific situations.

Ideally, we would calculate the entropies of $\bhy_{j}$, $\bhy_{j}|\bty_{1},...,\bty_{J}$ (conditioned on $\bty_{1},...,\bty_{J}$), and $\bhy_{SIj}$ which are correspondingly denoted by $H(\widehat{Y}_{j})$, $H(\widehat{Y}_{j}|\widetilde{Y}_{1},...,\widetilde{Y}_{J})$, and $H(\widehat{Y}_{SIj})$. We would compare $H(\widehat{Y}_{j})$ against $H(\widehat{Y}_{j}|\widetilde{Y}_{1},...,\widetilde{Y}_{J})\hspace{-2pt}+\hspace{-2pt}H(\widehat{Y}_{SIj})$ to choose between Intra-mode and Prior-mode. However, $\bty_{j}$ is not available at the encoder and it is only constructed at the decoder. Instead, we utilize $\by_{SIj}$ and project back to the dimension of $\by_{j}$ as a rough estimate of $\bty_{j}$ to determine the best mode.

In case the Prior-mode is selected, we wish to obtain the most efficient measurement $\by_{SIj}$ in terms of minimizing the total encoding rate, given by $H(\widehat{Y}_{j}|\widetilde{Y}_{j})\hspace{-2pt}+\hspace{-2pt}H(\widehat{Y}_{SIj})$. 
To this end, the measurement matrix $\mathbf{\Phi}_{SIj}$ is chosen as the solution of the following problem:
\begin{equation}\label{minRate}
    \mathbf{\Phi}_{SIj}=\argmin_{\mathbf{\Phi}_{SIji}}(H(\widehat{Y}_{j}|\widetilde{Y}_{ji})+H(\widehat{Y}_{SIji})),
\end{equation}
where $H(\widehat{Y}_{SIji})$ is the entropy of quantized $\widehat{\by}_{SIji}$ and $H(\widehat{Y}_{j}|\widetilde{Y}_{ji})$ is the entropy of $\bhy_{j}$ conditioned on the generated SI {$\bty_{ji}$}, through the corresponding projection $\mathbf{\Phi}_{SIji}$. To solve the problem in \eqref{minRate}, we can use a greedy approach, where different projected matrices are performed to find a projected matix that minimizes \eqref{minRate}.
\section{Experiment}
\label{Experiment}
\begin{figure*}[!t]
  \centering
    \subfigure[Object 58]{\label{figOBJ58SIFT}\includegraphics[width=0.75\textwidth]{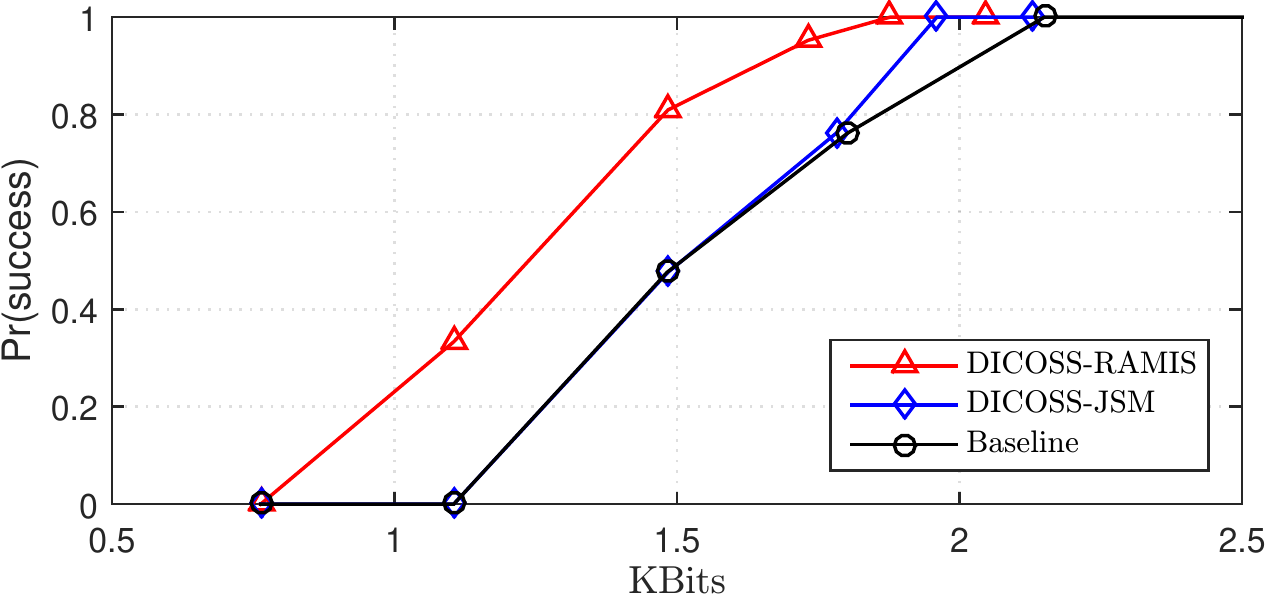}}
      \subfigure[Object 59]{\label{figOBJ59SIFT}\includegraphics[width=0.75\textwidth]{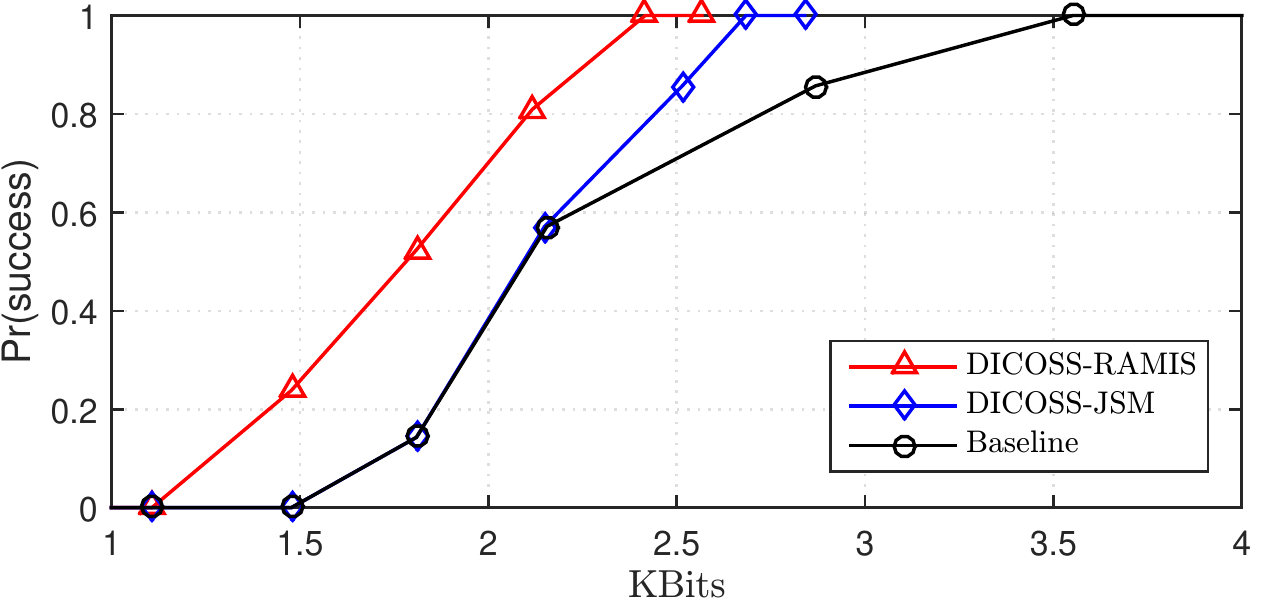}}
  \subfigure[Object 60]{\label{figOBJ60SIFT}\includegraphics[width=0.75\textwidth]{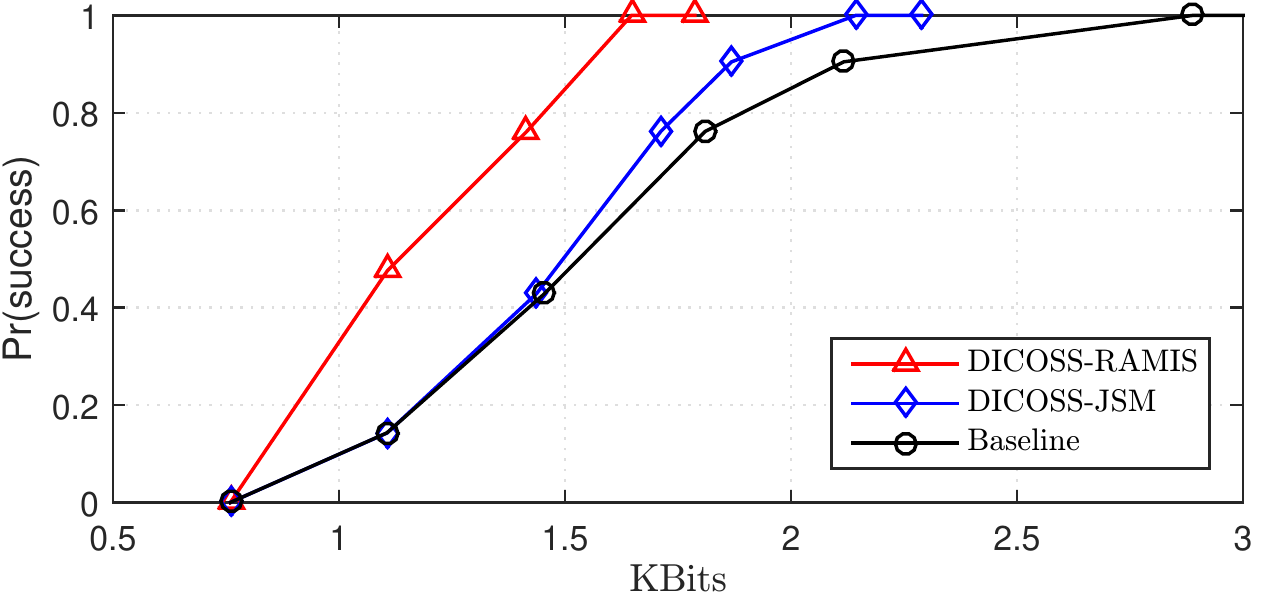}}
  \caption{Bits vs. reconstruction accuracy for the proposed DICOSS and Baseline for objects 58, 59, 60 in COIL-100.}
  \label{figOBJSIFT}
\end{figure*}
We consider sparse sources in the context of multiview object recognition, where a hierarchical vocabulary tree \cite{NisterCVPR06,NaikalFUSION10} is used for recognition and testing on a public object database, called COIL-100 \cite{Nene96}. COIL-100 contains multiview images of 100 small objects. SIFT \cite{DLowe99} features are extracted from the images of COIL-100. During the training stage, all features are clustered into a hierarchical vocabulary tree based on a hierarchical $k$-means algorithm \cite{NisterCVPR06}. The size of the tree depends on the value of $k$ and the number of hierarchies, e.g., if $k\hspace{-2pt}=\hspace{-2pt}10$ and 3 hierarchies, $n\hspace{-2pt}=\hspace{-2pt}1000$ vocabularies. In the testing phase, $J$ cameras acquire $J$ images of a given object, where all features of a given image $j$ are propagated down the tree to form a feature histogram vector, $\bx_{j}$ as in Sec. \ref{problem}. Hence, per query object, there are $J$ multiview histogram vectors, $\bx_{1},...,\bx_{J}\in \mathbb{R}^{n}$ to be used for recognition. Because of the small number of features in a single image, the histogram vector $\bx_{j}$ is sparse. 

We compare the coding efficiency of DICOSS against the DCS scheme in \cite{NaikalFUSION10,YangIEEE10}, we refer to the latter as Baseline  \cite{NaikalFUSION10} without prior information used in DCS, in terms of bits rather than only projection dimensions. More specifically, DICOSS sends each SI $\bhy_{SIj}$ individually using entropy coding plus additional $\bhy_{j}|\bty_{1},...,\bty_{J}$ using distributed coding whereas Baseline \cite{NaikalFUSION10} only quantizes and entropy codes each $\bhy_{j}$ separately. Obviously, Baseline can be considered as a special case of DICOSS when sending no at all prior information. This means that if a reliable estimate (Sec. \ref{AdaptiveRate}) is created, we would ensure the superior efficiency of DICOSS. 

In order to ensure that our experimental setup reflects a realistic scenario, we randomly select the 3 neighbor views of a given object over 72 views captured through 360 degrees in COIL-100 \cite{Nene96} as corresponding to 3 cameras. Specifically, the three neighbor views are assigned to $\bx_{1}$, $\bx_{2}$, $\bx_{3}$, respectively. For a fixed number of coding bits, $\mathrm{Pr(success)}$ is the number of times, in which the source $\bx_{j}$ is recovered as $\bhx_{j}$ with an error $||\bhx_{j}\hspace{-1pt}-\hspace{-1pt}\bx_{j}||_{2}/||\bx_{j}||_{2}\hspace{-2pt} \leq \hspace{-2pt}0.04$, divided by the total number of 100 trials (each trial considered different $\bx_{1},\bx_{2},\bx_{3}$). It is worth noting that the error is experimentally chosen according to how $\by_{j}$ is quantized and lossy transmitted to the decoder. In this experiment, each $\by_{j}$ (or $\by_{SIj}$) is uniformly quantized by 6 bits and decomposed into 6 $m_{j}$-bits length binary sequences which are in turn fed to the SW encoder.

Figure \ref{figOBJSIFT} presents the performance of DICOSS and Baseline per camera in terms of bits [Kbit] against the probability of successful reconstruction [\%] for objects 58, 59, 60 (Figs. \ref{figOBJ58SIFT}, \ref{figOBJ59SIFT}, \ref{figOBJ60SIFT}, respectively) in COIL-100 \cite{Nene96}. The DICOSS architecture (Fig. \ref{figDICOSS}) employs either JSM (Sec. \ref{JSM_Model}) or RAMIS (Sec. \ref{RAMIS_Model}) and the corresponding configurations are denoted as DICOSS-JSM and DICOSS-RAMIS. In general, the encoding rate required by DICOSS is significantly reduced compared to Baseline as shown in Fig. \ref{figOBJSIFT}. Particularly, the highest reduction of DICOSS-RAMIS is up to 43\% per camera at $\mathrm{Pr(success)}=1$ for object 60 in Fig. \ref{figOBJ60SIFT}. In addition, the encoding rate of DICOSS-RAMIS is systematically lower than that of DICOSS-JSM. These results reveal the potential of the proposed RAMIS in exploiting the correlations between the various signals versus JSM.
\section{Conclusion}
\label{conclusion}
This paper presented a novel approach to perform distributed sensing and encoding of multiview sources. The proposed DICOSS sent prior information to generate side information signals that help exploiting intra- and inter-source redundancies among multiple sparse sources. Moreover, we proposed a RAMIS algorithm that was integrated to DICOSS to improve the SI generation as well as the reconstruction of the multiview sources. The proposed DICOSS was shown to systematically yield bit-rate saving compared to Baseline without exploiting prior information. The experimental results showed improvements up to 43\% in terms of number of bits saved per camera for a given reconstruction accuracy.



\bibliographystyle{IEEEtran}
\bibliography{IEEEfull,IEEEabrv,bibliography}

\end{document}